\documentclass[runningheads]{llncs}
\usepackage{graphicx}
\usepackage{amsmath,amsfonts,amssymb}
\usepackage{pifont}
\usepackage{subcaption}
\usepackage{blindtext}
\usepackage{hyperref}
\usepackage{booktabs}
\usepackage{multirow}
\begin{document}
\title{Aligning Human Knowledge with Visual Concepts Towards Explainable Medical Image Classification}

\titlerunning{Aligning Human Knowledge with Visual Concepts}
%
\author{Yunhe Gao, Difei Gu, Mu Zhou, Dimitris Metaxas}

\authorrunning{Yunhe Gao, Difei Gu, Mu Zhou, Dimitris Metaxas}
%
\institute{Rutgers University}
\maketitle              
\begin{abstract}
Although explainability is essential in the clinical diagnosis, most deep learning models still function as black boxes without elucidating their decision-making process. In this study, we investigate the explainable model development that can mimic the decision-making process of human experts by fusing the domain knowledge of explicit diagnostic criteria. We introduce a simple yet effective framework, \textbf{Explicd}, towards \textbf{Exp}lainable \textbf{l}anguage-\textbf{i}nformed \textbf{c}riteria-based \textbf{d}iagnosis. Explicd initiates its process by querying domain knowledge from either large language models (LLMs) or human experts to establish diagnostic criteria across various concept axes (e.g., color, shape, texture, or specific patterns of diseases). By leveraging a pretrained vision-language model, Explicd injects these criteria into the embedding space as knowledge anchors, thereby facilitating the learning of corresponding visual concepts within medical images. The final diagnostic outcome is determined based on the similarity scores between the encoded visual concepts and the textual criteria embeddings. Through extensive evaluation of five medical image classification benchmarks, Explicd has demonstrated its inherent explainability and extends to improve classification performance compared to traditional black-box models. Code is available at \url{https://github.com/yhygao/Explicd}.

\end{abstract}
\section{Introduction}

\begin{figure}[t]
    \centering
    \includegraphics[width=1.0\textwidth]{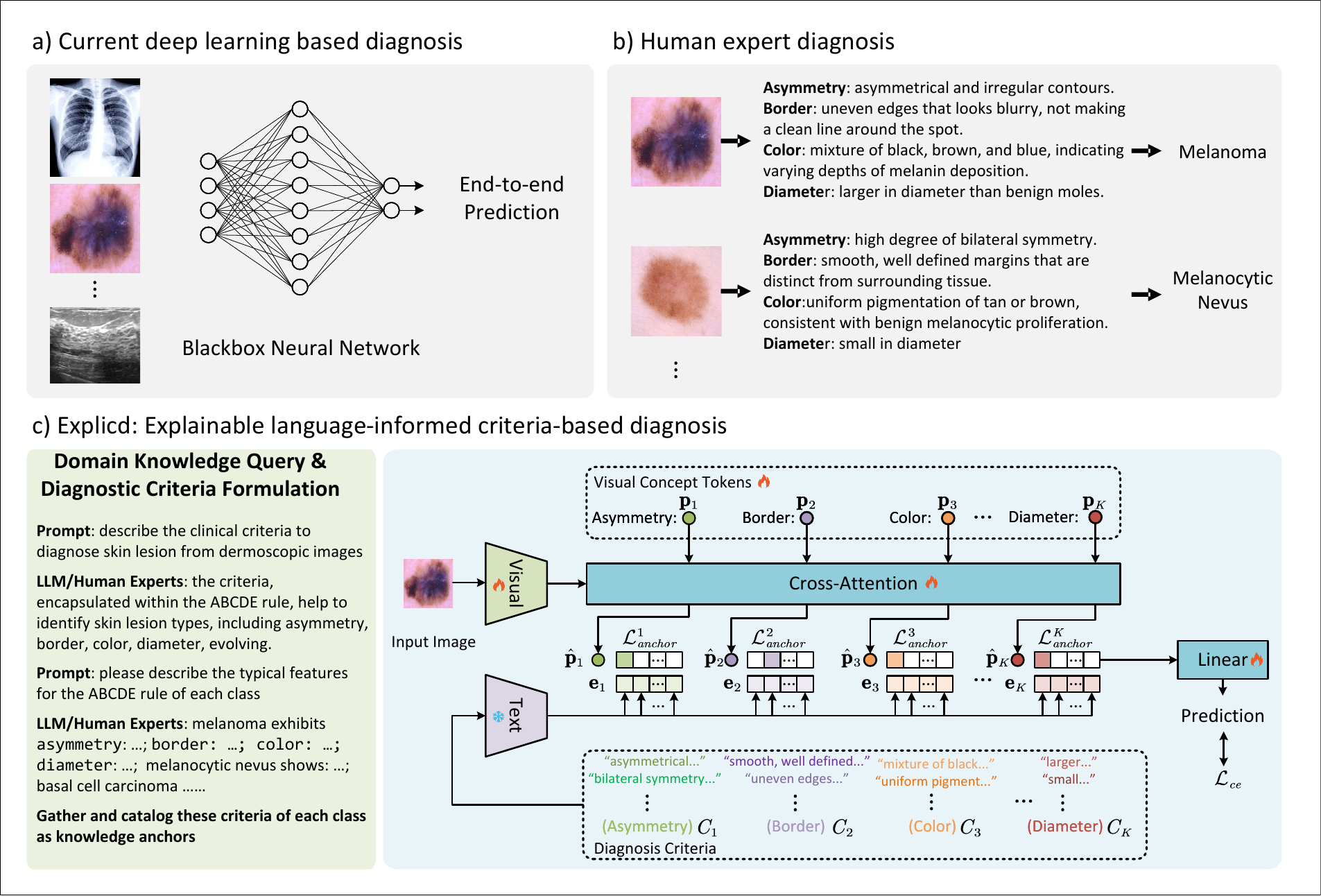}
    \caption{(a) Current state-of-the-art deep learning models often function as black boxes, offering predictions without revealing insights into their decision-making processes. (b) Depiction of the decision-making process by human experts for skin lesions, grounded in domain-specific knowledge and precise criteria, facilitates explainable diagnoses. (c) Overview of our Explicd framework: Domain knowledge is queried from LLMs or human experts across criteria axes. Explicd then aligns encoded visual concepts with textual knowledge anchors, facilitating the learning of visual concepts. The final diagnostic prediction is made based on the alignment scores between visual and textual concepts with a linear function.}
    \label{fig:ai_vs_human}
\end{figure}

The advent of deep learning~\cite{he2016deep,dosovitskiy2020image} has profoundly transformed the field of medical image analysis~\cite{li2014medical,cai2020review,kim2022transfer,gao2021utnet} in lowering diagnostic costs~\cite{khanna2022economics,gao2019focusnet,gao2021focusnetv2} and improving diagnostic accuracy~\cite{shen2015multi,aggarwal2021diagnostic}. Despite these advancements, state-of-the-art deep neural networks often operate as black boxes. While they can achieve high performance in end-to-end image classification, they fail to provide the explainable rationale behind the decision-making process. This lack of transparency compromises the trust and validation by healthcare professionals, leading to potential errors and resistance of integrating AI-derived insights into clinical settings~\cite{cutillo2020machine}. Unlike these black-box AI models, as illustrated in Fig. \ref{fig:ai_vs_human} (B), human experts make diagnoses by meticulously analyzing key image features from color, shape, size, or specific patterns regarding the disease symptom, staging, and outcome. In essence, human-expert decisions are grounded in a set of differential diagnostic criteria, enabling us to distinguish between various medical conditions with confidence and transparency. Developing explainable models that mirror the nuanced criteria defined by human knowledge is crucial to real-world AI clinical applications.

Incorporating human knowledge into the medical image assessment faces a long-standing challenge in the cross-modal data integration. The surge of vision-language models (VLMs)~\cite{radford2021learning,jia2021scaling,li2019visualbert} has focused on aligning image-text pairs in a unified representation space, opening up opportunities for the joint understanding of vision and language tasks. However, major general VLMs~\cite{radford2021learning} are not trained on medical data. They often fall short when dealing with medical image-text pairs due to a significant shift in data distribution~\cite{zhang2024data}. Although efforts have been placed on VLMs~\cite{boecking2022making,zhang2023large} pretrained on biomedical data, their performance still lags behind task-specific models in a broad range of medical image analysis tasks~\cite{windsor2023vision,gao2024training}. This is because they often struggle to align the associations between nuanced task-specific texts and the corresponding visual features. This inherent gap necessitates the development of better fine-grained alignment methods.

In this paper, we present \textbf{Explicd}, a simple yet effective framework for \textbf{Exp}lainable \textbf{l}anguage-\textbf{i}nformed \textbf{c}riteria-based \textbf{d}iagnosis. Typically, medical image classification benchmarks provide only the images and corresponding labels, largely omitting the detailed diagnostic information. Explicd addresses this gap by querying domain knowledge from either large language models (LLMs) like GPT-4~\cite{achiam2023gpt} or directly from human experts to formulate comprehensive diagnostic criteria, including key visual aspects such as color, shape, texture, or specific patterns associated with the classification task. We catalog the characteristics of each class based on these criteria axes. Within Explicd, these criteria are embedded as knowledge anchors using a VLM's text encoder. Moreover, we use a set of visual concept tokens to encode visual concepts from images along these criteria axes. An intermediate criteria anchor contrastive loss encourages high similarity between the encoded visual concepts and the corresponding positive knowledge anchors. Finally, a linear layer predicts the final diagnosis class by integrating the alignment scores from all criteria. Our contributions are as follows:
\begin{itemize}
    \item We introduce Explicd as a simple yet effective framework for explainable language-informed, criteria-based diagnosis via vision-language models.
    \item Explicd utilizes domain knowledge from LLMs or human experts, defining diagnostic criteria for each class across specific concept axes.
    \item We propose a visual concept learning module alongside a criteria anchor contrastive loss to align fine-grained visual features with diagnostic criteria.
    \item Explicd demonstrates superior interpretability and performance compared to traditional black-box models on five public benchmarks.
\end{itemize}

\section{Related Work.}

\noindent\textbf{Vision-language model.} VLMs learn joint representations from image-text pairs through contrastive learning. Despite the strong generalizability demonstrated by pioneering works~\cite{radford2021learning,jia2021scaling}, applying VLMs to the biomedical domain is challenging due to the distribution shift and domain-specific vocabulary. To mitigate this, biomedical VLMs like BioViL~\cite{boecking2022making} and BiomedCLIP~\cite{zhang2023large} have been pretrained on biomedical data like radiology reports and PubMed articles and outperform general VLMs on some biomedical tasks, but still fall short compared to task-specific models~\cite{windsor2023vision}. In this study, we propose a knowledge-based, fine-grained alignment method that adapts biomedical VLMs to specific diagnostic criteria, bridging the performance gap with task-specific models in medical image classification.

\noindent\textbf{Explainable model.}
Explainable AI models can be categorized into post-hoc and self-interpretable methods with a goal of making decision processes understandable~\cite{kakkad2023survey}. Post-hoc methods (e.g., Grad-CAM~\cite{selvaraju2017grad}) analyze the trained model to identify informative features, offering flexibility for pre-trained models, but they may not accurately reflect the model's true reasoning process~\cite{yang2023language}. On the other hand, self-interpretable methods design the explainability architecture directly inside the model. Concept bottleneck model (CBM)~\cite{koh2020concept} is a representative work that predicts predefined concepts to enable transparent decision-making, but it requires time-consuming attribute annotation. Our approach aligns with self-interpretable models but leverages large language models (LLMs) to bypass the annotation of concepts. LaBo~\cite{yang2023language} is a state-of-the-art self-interpretable model using visual-language concept scores, but relies heavily on the quality of pretrained VLM alignment. Our method enhances explainability by specifying diagnostic criteria axes and proposing visual concept learning for better alignment, ensuring learned concepts are strongly related to human experts' diagnostic criteria.

\section{Method}

Fig. \ref{fig:ai_vs_human} (C) illustrates the proposed Explicd framework. In this section, we present details on how Explicd queries knowledge as diagnostic criteria, aligns visual features with these criteria, and makes explainable classifications.

\subsection{Domain Knowledge Query \& Diagnostic Criteria Formulation}
Disease diagnosis usually centers around various criteria axes describing distinctive characteristics across clinical classes~\cite{liberman2002breast,turkbey2019prostate}. Drawing from inspiration, we first query domain knowledge from LLMs or consult human experts and formulate them into textual diagnosis criteria. Consider a set of training image-label pairs $D=\{(x, y)\}$, where $x$ is the image and $y\in \mathcal{Y}$ is a label from a set of $N$ classes. Specifically, we categorize the diagnosis criteria along $K$ disentangled \textit{criteria axes} specified by language depending on the task $\{C_i\}_{i=1}^K$. For instance, in the case of skin lesions, the criteria axes include \texttt{asymmetry, border, color, diameter, texture, pattern}. Subsequently, we query detailed knowledge on the typical characteristics for each class along each criteria axis $C_i = \{c_i^1, \dots, c_i^{n_i}\}$, where $1<n_i\leq N$ denotes the number of possible options within a particular criteria axis. Take the \texttt{color} of skin lesion as an example, potential options could range from `a mixture of black brown and blue' for melanoma, `uniform pigmentation of tan or brown' for melanocytic nevus, among others. Notably, the quantity of typical characteristics for each criteria axis, $n_i$, may be less than the number of classes $N$, as different classes might exhibit identical characteristics for certain criteria axes; e.g., various types of benign skin lesions could all present symmetry. Additionally, the ground truth label for each diagnostic criterion is recorded, i.e. for each criteria axis, the associated class and characteristic options are marked as positive, whereas all other combinations are considered negative.

\subsection{Visual Concept Learning}
After collecting the textual form diagnostic criteria, we aim to align the visual features with these textual human knowledge. In particular, we propose a lightweight visual concept learning module for aligning the fine-grained visual concepts and the nuanced textual criteria. Specifically, given a pretrained vision-language model with visual encoder $\mathcal{V}$ and text encoder $\mathcal{T}$, we first encode the queried diagnostic criteria into criteria anchor embeddings, $\{\mathbf{e}_i = \mathcal{T}(C_i)\}_{i=1}^K$, where $\mathbf{e}_i \in \mathcal{R}^{n_i\times d}$, and $d$ is the dimension of the embedded token. These criteria embeddings act as a sparse representation of human knowledge, serving as anchors to facilitate the learning of visual concepts. 

To capture visual concepts effectively, our visual concept learning module employs a set of $K$ learnable visual concept tokens $\mathbf{p}\in \mathcal{R}^{K\times d}$, with each token designated to represent one of the $K$ criteria axes.  For a given image $x$ and its feature map $\mathcal{V}(x)$, the concept encoding process is formalized as follows: 
\begin{equation}
    \hat{\mathbf{p}} = \text{cross-attention}(\mathbf{p}, \mathcal{V}(x), \mathcal{V}(x)),
\end{equation}
where $\mathbf{p}$ is the query and $\mathcal{V}(x)$ serves as the key and value of the cross-attention layer. The visual concept tokens $\mathbf{p}$ interact with the image feature map, thereby encoding the relevant visual concepts associated with specific criteria axes into $\hat{\mathbf{p}}\in \mathcal{R}^{K\times d}$.

The learning of visual concepts is facilitated by a contrastive loss. For each criteria axis, the aggregated visual concept token $\hat{\mathbf{p}}_i$ is compared against all characteristic embeddings $\mathbf{e}_i$, calculating a similarity score. The criteria anchor contrastive loss is as follows:
\begin{equation}
\mathcal{L}_{anchor}^{i}(\hat{\mathbf{p}}_i, \mathbf{e}_i) = -\log \frac{\exp(sim(\hat{\mathbf{p}}_i, \mathbf{e}_i^{\text{positive}} / \tau)}{\sum_{j=1}^{n_i} \exp(sim(\hat{\mathbf{p}}_i, \mathbf{e}_i^j) / \tau)}
\end{equation}
where $\tau$ denotes the temperature parameter that adjusts the softness of the softmax distribution and we use dot product as the similarity function. The criteria anchor contrastive loss aims to increase the similarity between the encoded visual concept token $\hat{\mathbf{p}}_i$ and the positive criteria anchor embeddings while 
decreasing its similarity with the embeddings of negative characteristics, ensuring a more discriminative learning of visual concepts along each diagnostic criteria axis.

\subsection{Explainable classification}
The above knowledge anchor loss $\mathcal{L}_{anchor}$ enables the alignment of encoded visual concepts with the corresponding characteristic options along each diagnostic criteria axis. Intuitively, the similarity scores between the encoded visual concept token $\mathbf{p_i}$ and the diagnostic criteria anchor $\mathbf{e}_i$ indicate the model's assessments for each diagnostic criterion. Mirroring the approach of human experts, who make their final diagnosis on the evaluations across multiple criteria, we use a linear layer to make prediction of the final class by integrating the alignment scores from all $K$ criteria axes. 
\begin{equation}
    \hat{y} = W (\text{concat}(sim(\hat{\mathbf{p}}_1, \mathbf{e}_1), \dots, sim(\hat{\mathbf{p}}_K, \mathbf{e}_K)))^{\intercal},
\end{equation}
where $\text{concat}(,)$ represents the concatenation operation and $W$ is the weights in the linear layer that inherently reflect the significance of each diagnostic criterion's contribution towards the overall class prediction.

During the training phase, we optimize a joint objective that includes both the criteria anchor contrastive loss with cross-entropy loss for the final classification:
\begin{equation}
    \mathcal{L}_{total} = \mathcal{L}_{\text{ce}}(\hat{y}, y) + \frac{1}{K}\sum_{i=1}^{K}\mathcal{L}_{anchor}^i(\hat{\mathbf{p}}_i, \mathbf{e}_i)
\end{equation}

The embeddings of textual criteria anchors are precomputed and stored, ensuring that the training and inference overhead introduced by the additional components are negligible.

\section{Experiments}
\subsection{Experimental Setup}
We evaluate our method on five publicly available medical image classification benchmarks, which cover a diverse range of medical targets and modalities. 

\noindent \textbf{Dataset:} \textbf{ISIC2018}~\cite{tschandl2018ham10000} contains 10,015 dermoscopic images with seven skin lesion categories for skin cancer classification. \textbf{NCT-CRC-HE}(NCT)~\cite{kather2018100} includes 100,000 patch-based histological images of human colorectal cancer for training and 7180 patches for validation, with nine tissue classes for classification. \textbf{IDRiD}~\cite{porwal2018indian} consists of 516 retinal fundus images annotated with 5 severity level grading of diabetic retinopathy. \textbf{BUSI}~\cite{al2019deep} dataset contains 780 ultrasound images of breast masses categorized into normal, benign, and malignant classes for breast cancer classification. \textbf{MIMIC-CXR}~\cite{johnson2019mimic}
contains 377,100 chest X-ray images. Cardiomegaly (CM) and Edema are used for binary classification. 

\noindent \textbf{Baselines.} We compare our method with several baselines, including (1) \textbf{VLMs zero-shot}: We apply the general VLM CLIP~\cite{radford2021learning} and biomedical VLMs BioViL~\cite{boecking2022making} and BiomedCLIP~\cite{zhang2023large} in a zero-shot setting for classification; (2) \textbf{Supervised black-box models}: We fine-tune ImageNet-pretrained ResNet50~\cite{he2016deep} and ViT-Base~\cite{dosovitskiy2020image} on the classification benchmarks; and (3) \textbf{LaBo}: a state-of-the-art explainable model with concept bottleneck.

\noindent \textbf{Implementation Details.} We prompt \href{https://platform.openai.com/docs/models/gpt-4-and-gpt-4-turbo}{GPT-4} to query domain knowledge and diagnostic criteria. We use the official implementation and pretrained weights of \href{https://huggingface.co/docs/transformers/model_doc/clip}{CLIP ViT-Base}, \href{https://huggingface.co/microsoft/BiomedVLP-BioViL-T}{BioViL} and \href{https://huggingface.co/microsoft/BiomedCLIP-PubMedBERT_256-vit_base_patch16_224}{BiomedCLIP}. Our Explicd and LaBo are implemented based on BioViL-specialized for MIMIC-CXR dataset, while using BiomedCLIP for all other datasets. The fine-tuning of Explicd involves optimizing visual encoder, visual concept learning module and the final linear layer with AdamW optimizer, while keeping the text encoder fixed. All experiments are conducted using PyTorch with Nvidia A6000 GPUs.

\begin{table}[t]
    \centering
    \scriptsize
    \caption{Performance comparison across five benchmarks. Balanced accuracy is reported for CM and edema in MIMIC-CXR due to class imbalance; accuracy is reported for the other datasets.}
    \setlength{\tabcolsep}{2.5mm}{
    \begin{tabular}{c|c|c|c|c|c|c|c}
    \toprule
    Setting & Model & ISIC2018 & NCT & IDRiD & BUSI & CM & Edema  \\\hline
    \multirow{3}{*}{Zero-shot} & CLIP & 11.6 & 9.9 &  31.1 & 30.8 & 49.5 & 51.4\\
    & BioViL & 8.5 & 7.7 & 26.2 & 30.8 & 70.8 & 76.9\\
    & BiomedCLIP & 21.2 & 35.3 & 37.9 & 37.2 & 69.3 & 77.1 \\\hline
    \multirow{2}{*}{Black-box} & ResNet50 & 82.6 & 93.4 & 53.4 & 84.6 & 79.7 & 77.4 \\
    & ViT-Base & 89.0 & 94.4  & 57.3 & 88.5 & 79.2 & 80.9\\\hline
    \multirow{2}{*}{Explainable} & LaBo & 80.9 & 90.2 & 48.4 & 75.8 & 73.5 & 74.2\\
    & Explicd (ours) & \textbf{90.0} & \textbf{95.1} & \textbf{58.5} & \textbf{89.7} & \textbf{81.8} & \textbf{85.7}\\\bottomrule
    
    \end{tabular}}
    \label{tab:main_results}
\end{table}

\subsection{Main Results}

Our proposed Explicd model demonstrates superior performance compared to various baseline methods across five medical image classification benchmarks, as shown in Table \ref{tab:main_results}. The zero-shot performance of VLMs, including CLIP, BioViL, and BiomedCLIP, is generally poor, with CLIP performing close to random guessing across all datasets. This indicates that CLIP's visual-text alignment is not effective for complex medical diagnosis tasks as it is trained on general vision data. Although BioViL and BiomedCLIP perform much better on the MIMIC-CXR dataset (CM and Edema tasks), likely due to their pretraining data being largely based on chest X-ray radiology reports, their performance on other datasets remains much lower than supervised trained black-box models like ResNet50 and ViT-Base, suggesting their limited generalization ability.

Explicd effectively combines explainability with high-level classification performance, outperforming not only the explainable model LaBo but also black-box models across all datasets. LaBo's lower accuracy compared to the black-box models can be attributed to its reliance on well-aligned vision-language models, highlighting the challenge of maintaining high accuracy while providing strong explainability. In contrast, Explicd's superiority is due to the introduction of human knowledge and visual concept learning, which provides additional supervision for fine-grained alignment between visual features and diagnostic criteria. This strategy not only enhances fine explainability but also brings improvement in overall classification performance.

\subsection{Diagnostic Interpretation}

\begin{figure}[t]
     \centering
     \includegraphics[width=0.9\textwidth]{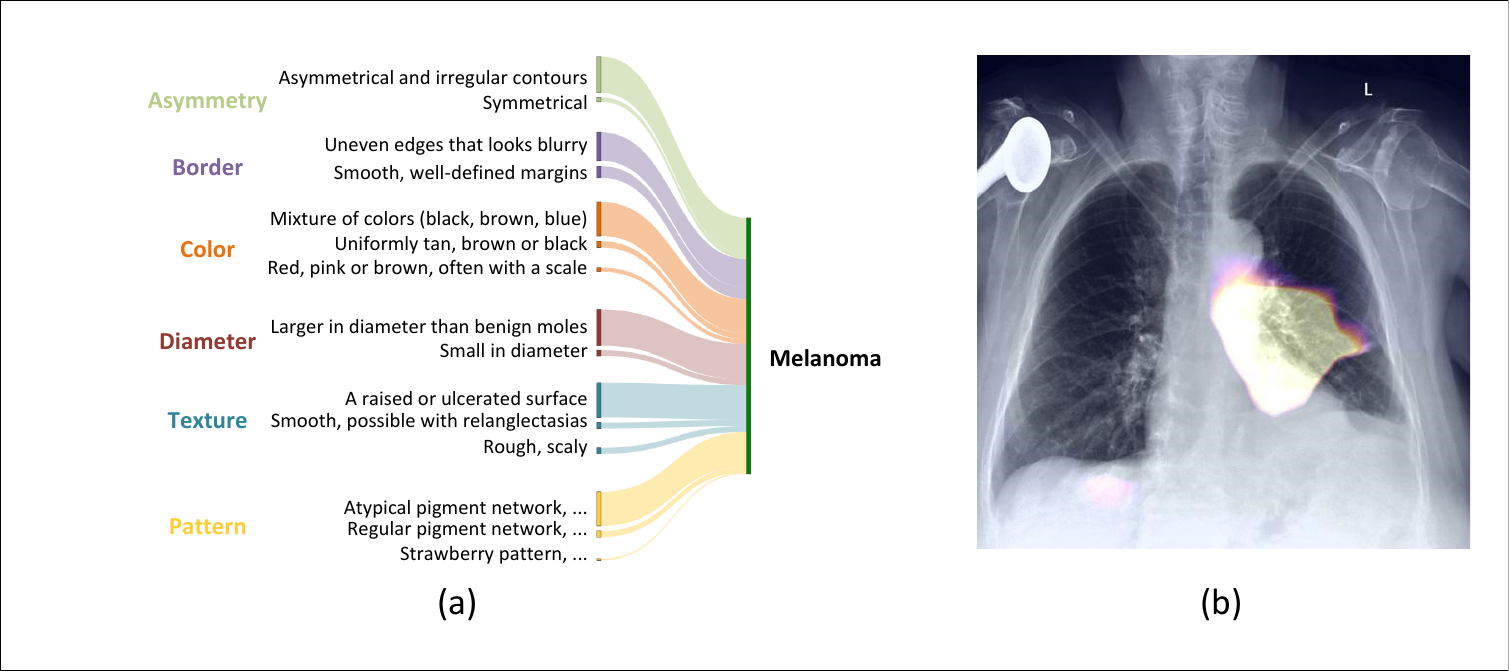}
     \caption{(a) Alignment scores measured using cosine similarity between the encoded visual concept tokens and diagnostic criteria along each axis for skin lesion classification. The width of the lines represents the strength of similarity, with wider lines indicating higher scores. (b) Heatmap visualization of the encoded visual concept tokens overlaid on the image feature maps for a case of cardiomegaly. Brighter regions indicate higher similarity scores, suggesting a stronger focus on these areas by the model.}
     \label{fig:interpretability}
 \end{figure}

A distinguishing design of Explicd is its appealing ability to interpret its decision-making process. Fig. \ref{fig:interpretability} (a) shows the alignment scores measured with cosine similarity between the encoded visual concept tokens and the embeddings of diagnostic criteria for skin lesions. The width of the lines indicates the strength of similarity, with larger widths representing higher similarity scores. We can see that Explicd can accurately predict the characteristics of each criteria axis, such as the presence of asymmetry, border irregularity, color variegation, and large diameter, which are key features in the diagnostic criteria we queried for melanoma diagnosis. The high similarity scores between the visual concepts and the corresponding diagnostic criteria demonstrate that Explicd has learned to identify and align these important visual features, leading to a correct final diagnosis of melanoma.

Furthermore, we visualize the heatmap of the average visual concept tokens with the image feature map of cardiomegaly in a chest X-ray image in Fig. \ref{fig:interpretability} (b). The brighter regions indicate higher similarity scores, suggesting that the model is focusing on these areas when making its prediction. Cardiomegaly is a medical condition characterized by an enlarged heart. In the heatmap, we can observe that Explicd correctly focuses its attention on the heart area, indicating that Explicd has learned to align human knowledge regarding the key visual features of cardiomegaly with the relevant visual concepts in the X-ray image. By providing the alignment scores on criteria and highlighting the most important regions contributing to its prediction, Explicd provides a transparent and interpretable decision-making process that can be easily understood and verified by medical experts.

\section{Discussion}

To address the lack of transparency and interpretability in current deep learning models for medical image analysis, we proposed Explicd, a comprehensive framework that integrates diagnostic criteria queried from LLMs, aligning visual concepts towards explainable classification. Explicd offers a novel means to understanding diseases along human-understandable criteria axes. Our extensive experiments highlight Explicd's superior performance over both traditional black-box approaches and existing explainable models, setting a new standard in both accuracy and interpretability. The clarity of Explicd’s decision-making process promises to bolster trust and facilitate the integration of AI in clinical diagnostics. Moving forward, we aim to expand the incorporation of broad human knowledge within our diagnostic criteria and refine the hierarchical representations of visual concepts, allowing for a more nuanced exploration of disease diagnosis and management.

\begin{credits}
\subsubsection{\ackname} This research has been partially funded by research grants to D. Metaxas through NSF: 2310966, 2235405, 2212301, 2003874, and FA9550-23-1-0417 and NIH 2R01HL127661.

\subsubsection{\discintname}
The authors have no competing interests to declare
that are relevant to the content of this article. 
\end{credits}

%
%
%
\bibliographystyle{splncs04}
\bibliography{reference.bib}

\end{document}